\documentclass{article}
\usepackage{ijcai17}
\usepackage{amsmath}
\usepackage{amsfonts}
\usepackage{algorithm}
\usepackage[noend]{algpseudocode}
\usepackage{csquotes}
\usepackage{graphicx}
\usepackage{color}
\usepackage{times}
\usepackage[utf8]{inputenc}
 
\makeatletter
\def\BState{\State\hskip-\ALG@thistlm}
\makeatother

\newlength\myindent
\title{Optimizing Gross Merchandise Volume via DNN-MAB Dynamic Ranking Paradigm} 
\author{Wentao Guo~~~Meng Zhao~~~Jinghe Hu~~~Weipeng P. Yan~~~Yan Yan\\ 
Business Growth BU., \\
Tower A, Beichen Century Center, 10\# Courtyard, \\
Beichen W. Rd., Chaoyang Dist., Beijing, China, 100101\\
\{guowentao,~zhaomeng1,~hujinghe,~paul.yan,~yan.yan\}@jd.com}

\begin{document}

\maketitle

\begin{abstract}
With the transition from people's traditional `brick-and-mortar' shopping to online mobile shopping patterns in web 2.0 \emph{era}, the recommender system plays a critical role in E-Commerce and E-Retails.  This is especially true when designing this system for more than $\mathbf{236~million}$ active users.  Ranking strategy, the key module of the recommender system, needs to be precise, accurate, and responsive for estimating customers' intents.  We propose a dynamic ranking paradigm, named as DNN-MAB, that is composed of a pairwise deep neural network (DNN) \emph{pre}-ranker connecting a revised multi-armed bandit (MAB) dynamic \emph{post}-ranker.  
By taking into account of explicit and implicit user feedbacks such as impressions, clicks, conversions, etc. DNN-MAB is able to adjust DNN \emph{pre}-ranking scores to assist customers locating items they are interested in most so that they can converge quickly and frequently.  To the best of our knowledge, frameworks like DNN-MAB have not been discussed in the previous literature to either E-Commerce or machine learning audiences.  In practice,  DNN-MAB has been deployed to production and it easily outperforms against other state-of-the-art models by significantly lifting the gross merchandise volume (GMV) which is the objective metrics at JD.   
\end{abstract}

\section{INTRODUCTION}
Effectively generating the right list of items for customers is the key to the success of E-Commerce websites and applications.  How fast and how frequent the customers could converge (place orders) decide if the E-Commerce company would thrive.  The recommender system, an information filtering, rating, and recommending engine that assists users to reach a small group of items that describe and satisfy their purchasing needs in real time, is emerged for solving this challenge.  

\begin{figure}
\includegraphics[height=6.25cm]{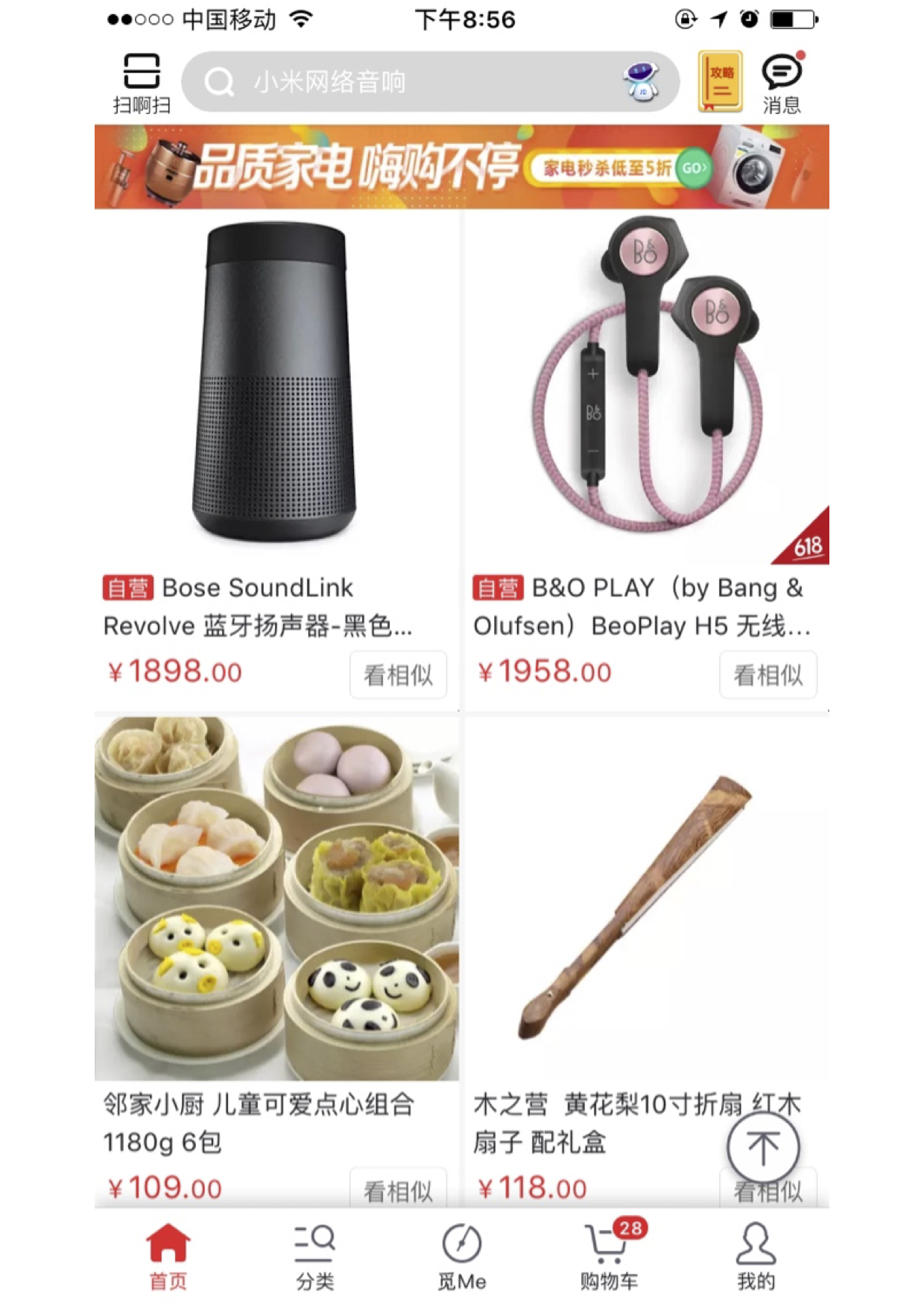}
\includegraphics[height=6.25cm]{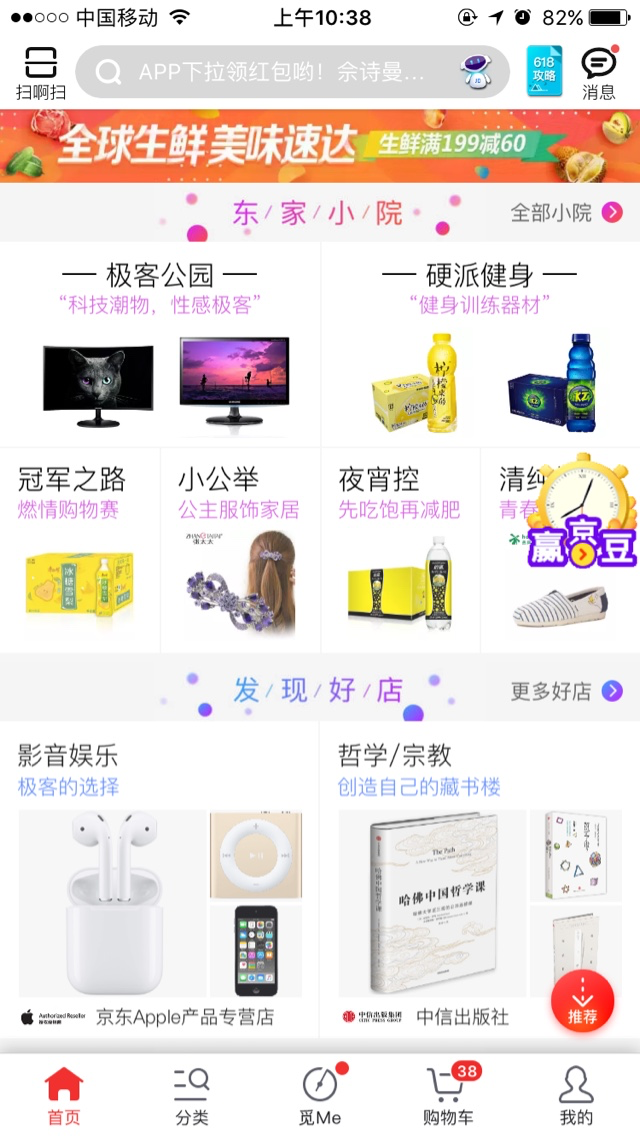}
\caption{two types of mobile page layouts of item rankings}
\label{graph:example-layout}
\end{figure}

Founded in 2004, JD has quickly ascended into one of the most popular E-Commerce websites in China.  Customers come to JD to discover, browse and purchase items sold by itself as well as over hundreds of thousands other government certificated E-Retailers.  Everyday tens of millions of users generate billions of querying requests, place tens of millions of orders.  At the same time, the statistics shows that in average the active users visit only limited number of items from specific positions like the front page, topic driven recommending sections in JD's \emph{app}.  (Fig.\ref{graph:example-layout} presents two type of layout).  Therefore, despite the huge amount of serving loads, the number of ranked items presented to users are actually small, the recommender system at JD must provide robust, agile and accurate recommendation service to make sure conversion happens frequently in every user's item browsing experience.  

The ranking module decides how the final list of items from the retrieval should be generated and positioned so that the items interest the customers most are placed first.  As a result, the ranking problem is always the centric issue of the recommender system.  In the use case of E-commerce recommendations, people mine each customer's information including search-history, clicks, orders, etc. to model customers' purchasing intent across the whole platform.    

Traditional ranking modules in recommender systems are not as efficient regarding with learning the users' intent and behavioral feebacks while they browse the ranked items.  This is due to the fact that most of the ranking results in the E-Commerce \emph{app}s come in a fashion of the waterflow streaming.  Traditional ranking ideas usually have difficulties to incorporate the users' real time feedbacks so that the results are either not not precise enough or in the extreme cases not legitimate anymore (\emph{e.g.} the static ranking results still promote items that users just clicked or placed orders on, based on recent behavioral histories).  In this paper, we propose an innovative dynamic ranking paradigm.  By combining the deep learning model and the multi-armed dandit algorithm together, this framework is capable of learning customers' real time feedbacks so to change the ranking results for reflectng users' current purchasing intent and improving the overall recommender system performance.  

\subsection{Contributions}
This paper has two primary contributions in research and industry: 
\begin{enumerate}
\item an innovative ranking paradigm for solving the dynamic ranking problem by combining the pairwise deep neural network and multi-armed bandit algorithms;
\item a revised Thompson sampling algorithm with the brand new initialization strategy under the production use case that enables the customers' fast convergence. 
\end{enumerate}
\subsection{Organization}
The rest of this paper is organized as follows: Sec.\ref{sec:formulation} briefly describes the recommender system utilized in JD and discusses the ranking module in detail: \ref{subsec:deep_learning} \emph{learning-to-rank} DNN module as the \emph{pre}-ranker and \ref{subsec:multi-armed-bandit} MAB Thompson sampling as the \emph{post}-ranker; 
Sec.\ref{sec:experiment} first discusses a simple case study which simulates the real world for testing how Thompson sampling works comparing with other popular MAB algorithms and then reports the proposed DNN-MAB performance regarding with different metrics; Sec.\ref{sec:related} discusses some prominent related studies regarding with recommender systems, \emph{learning-to-rank} via deep neural network models and multi-armed bandit models; we summarize our work and point out some future directions in Sec.\ref{sec:conclusion}. 

\section{FORMULATION}
\label{sec:formulation}
\subsection{System design}
\label{subsec:system_outline}
\begin{figure}
\centering
\includegraphics[width=9cm,height=4.3cm]{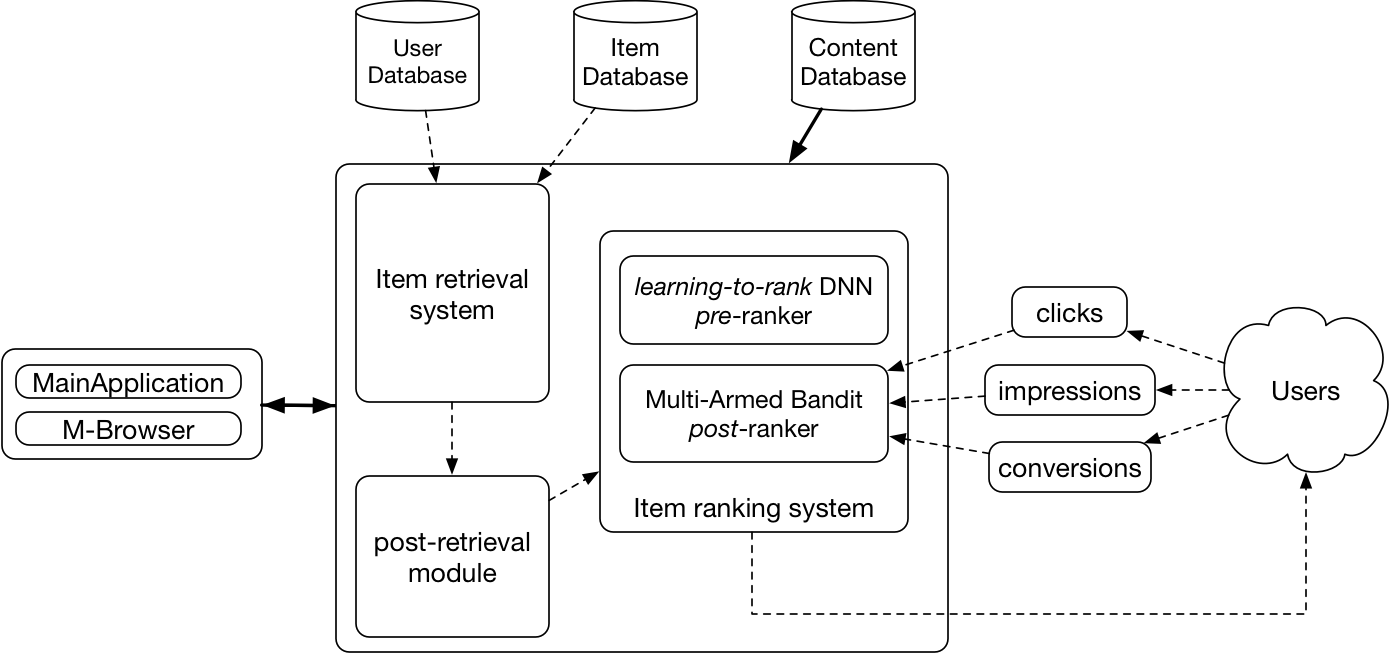}
\caption{the recommender system flowchart}
\label{graph:recommender_system}
\end{figure}

Our recommender system in Fig.\ref{graph:recommender_system} includes three main function modules: the item-retrieval module, the item post-retrieval module, and the ranking module.  Users typically trigger the recommender system via directing from the mobile application entry point or from our mobile version websites.  After the system indicates users' identities, the recommender issues a query request to the user database and the item database to fetch other information regarding with user profiles such like: gender, geo location, purchasing history, price sensitivity etc.  This piece of information combining with the items that the users have either recently clicked or put in carts are collected and serve as the input of the retrieval system.  Next, the retrieval system selects a large pool of candidate items that are related to the input. 

The motivation for the post-retrieval module is to filter out items that are not suitable for recommending, including items that users have already purchased, items containing sensitive contents, and other disqualified items etc.

The ranking module compares all candidate items provided from the post-retrieval module and generate the final top-$K$ item sublist.  Such list of items should be optimized so that the items that users interested in most should be placed at top positions.  Generally this is achieved by sophisticated machine learning models such as Factorization Machines \cite{rendle2010factorization}, pairwise \emph{learning-to-rank} \cite{burges2005learning,severyn2015learning}, listwise \emph{learning-to-rank} \cite{cao2007learning}, etc.  
In our use case, the ranking module is implemented by a pairwise loss \emph{learning-to-rank} DNN and a revised MAB Thompson sampling.

\subsection{Input data}
We now formally propose the ranking problem. Assuming that each sample is indicated as a $m$-dimentional feature vector $\boldsymbol{\xi} \in \mathbb{R}^m$ coming from a certain category $c \in \mathbb{R}^1$, with their $gmv \in \mathbb{R}^1$.  By given a set of items $\mathcal{T}_N~=~\{\boldsymbol{\xi}\}_N$ and a template triggering item $\boldsymbol{\xi}_\mathrm{tem}$, a subset of $K$ items are selected from $\mathcal{T}_N$ and ordered in a particular way $\{\boldsymbol{\xi}_1, ~\boldsymbol{\xi}_2,~\ldots,~\boldsymbol{\xi}_{K}\}$ so that the $dcg$ of the sublist's GMV is optimized: 

\begin{eqnarray}
dcg_K~=~\sum^K_{i=1}\cfrac{gmv_i\mathcal{I}(\boldsymbol{\xi}_i)}{\log_2 (i + 1)}
\label{eqn:def-ndcg}
\end{eqnarray}

Here, $\mathcal{I}(\boldsymbol{\xi})$ is the indicator function such that:
\begin{eqnarray}
\mathcal{I}(\boldsymbol{\xi})\left\{
  \begin{array}{rc}
    1 & \boldsymbol{\xi}\mathrm{~ordered~by~user~~~~~~} \\
    0 & \boldsymbol{\xi}\mathrm{~not~ordered~by~user} \\
  \end{array}
\right.
\end{eqnarray}

\subsection{\emph{learning-to-rank} DNN \emph{pre}-ranker}
\label{subsec:deep_learning}
\begin{figure}
\centering
\includegraphics[width=8.8cm,height=4.3cm]{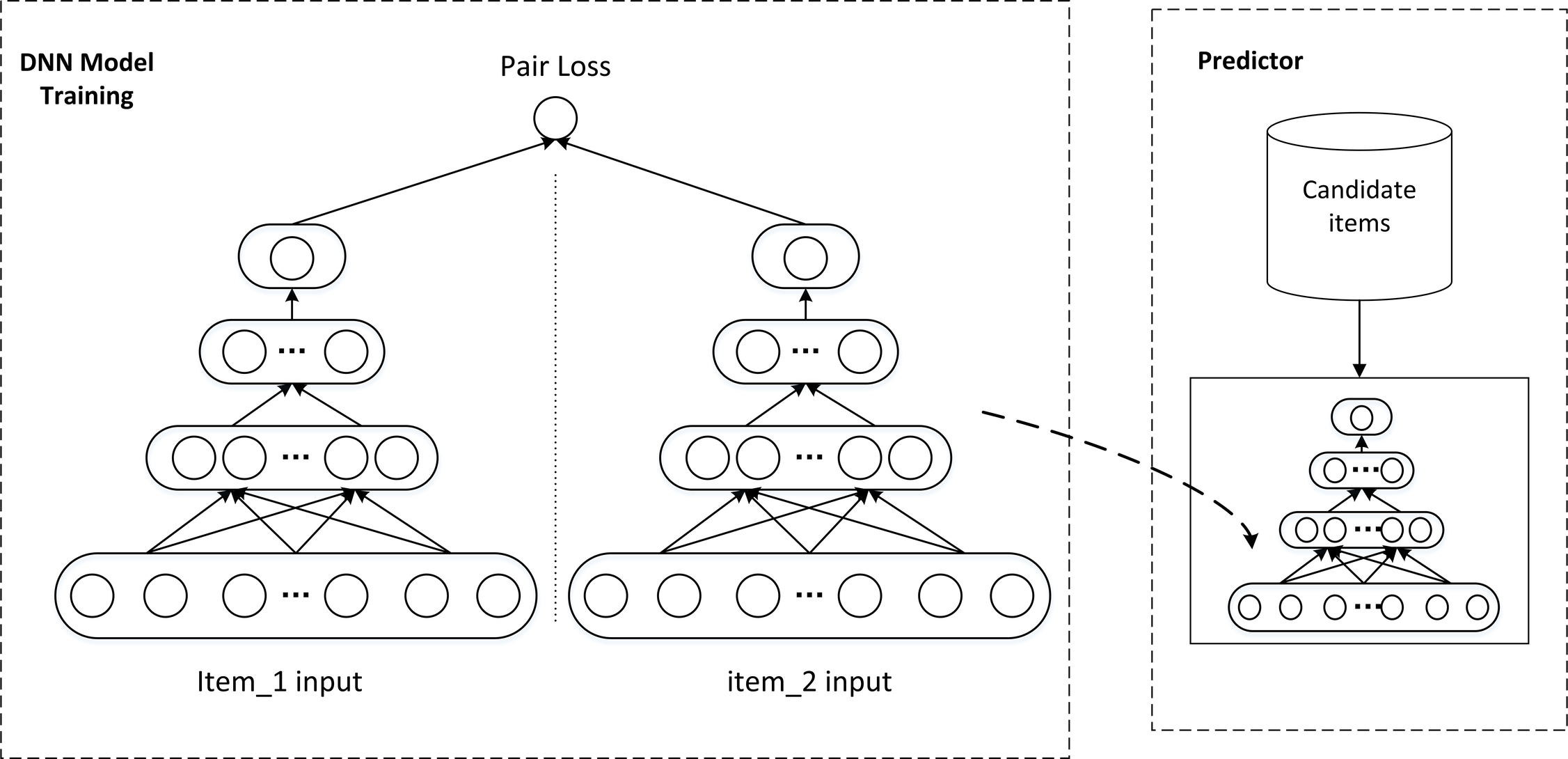}
\caption{the \emph{learning-to-rank} DNN pre-ranker}
\label{graph:dnn}
\end{figure}
The \emph{learning-to-rank} DNN model is to compute the pairwise loss for ranking different items based upon the label information of whether users have ordered/ not-ordered certain items.  The simplified model structure is shown as Fig.\ref{graph:dnn}.  It is implemented by two miorring $5$-layers DNNs: one input layer which takes item features that are sparse $m$-dimensional vectors, three fully-connected layers that generalize item features $\mathbf{x}$, and one output layer $\mathbf{w}$ \cite{basak2007support,cherkassky2004practical} which outputs $y$s serving as the \emph{pre}-ranker regression scores.  Noting that both DNNs share the same set of parameters.  

\begin{eqnarray}
\hskip-.3cm\mathcal{L}_H = {\scriptstyle\sum}^N_{i = 1}\lambda(\mathbf{x}_{1i},\mathbf{x}_{2i})\max(0,m - (y_{1i} - y_{2i})(t_{1i} - t_{2i}))\nonumber\\
y = ~\mathbf{x}^T\mathbf{w}~~~~~~~~~~~~~~~~~~~~~~~~~~~~~~~~~~~~~~~~~~~~~~~~~~~~~~~~~~~~~~~~~~~~~~~~
\label{eqn:svr-loss}
\end{eqnarray}

The loss function $\mathcal{L}_H$ in Eq.\ref{eqn:svr-loss} is inspired by SVM-\emph{rank} \cite{elisseeff2001kernel}, where $\mathbf{x}_{1i}$ and $\mathbf{x}_{2i}$ are the pair of item features labeled as $t_{1i}$ and $t_{2i}$.  The label $t$ is valued as either $0$ (negative) or $1$ (positive).  Each pair is generated in such a way that only one out of two items is from the positive class: $t_{1i} + t_{2i} = 1$.  $m$ is the tunning parameter representing the classification margin making sure that the better separability between two classes is preferred.  $\lambda(\cdot,\cdot)$ is the pairwise weighting function that emphasizes the losses from pairs of greater $gmv$ values.   

In the training phase, $N$ positive and negative item pairs input to both sides of the DNN models shown as the of Fig.\ref{graph:dnn} (L), since both DNNs are sharing the same parameters, \emph{learning-to-rank} DNNs learn from the pair-labeling difference and aim to find the scoring scheme that correctly classifies items with the largest margin.  

In the testing phase, each item from $\mathcal{T}_N$ passes through the predictor (Fig.\ref{graph:dnn} (R)), and is evaluated and scored by \emph{learning-to-rank} DNN.  The \emph{pre}-ranker scores $y_i$s are then served as the candidate scores for the MAB \emph{post}-ranker.  

\subsection{Multi-armed bandit \emph{post}-ranker}
\label{subsec:multi-armed-bandit}
The reasons for designing the MAB \emph{post}-ranker are mainly the following three:
\begin{enumerate}
\item the real-time `click' and `order' types of labels represent the user's current purchasing intent out of many other recent intents and it should be emphasized in rankings;
\item the real-time `no-action' labels indicate item categories that the user is not currently interested in and they should be de-emphasized in rankings;
\item users tend to click items under the same category and place orders by comparing them over different attributes.
\end{enumerate}
The first and second reasons are well discussed in \cite{radlinski2008learning} by stating that static rankings contain lots of redundant results.  The MAB \emph{post}-ranker is to emphasize items that users are potentially interested in by referencing other items clicked; de-emphasize items that users are intentionally ignored, meanwhile exploring items from different categories to diversify the ranking results.  

We follow the problem settings of the contextual multi-armed bandit problem in \cite{langford2008epoch}, and define the problem as follows: 
\subsubsection{Definition 2.4.1 (Contextual bandit in DNN-MAB)}
Given $M$ arms $\mathcal{C}_M = \{c_1,~c_2,~\ldots,~c_M\}$, and a set of items $\mathcal{T}_N = \{\mathbf{x}\}_N$ scored from \emph{learning-to-rank} DNN that each item belongs to exactly one out of $M$ arms.  The player needs to pull one arm $c_i \in \mathcal{C}_M$ at each round $i$, so that the item $\mathbf{x}_i$ from that arm is picked up and placed at position $i$.  The reward at round $i$ is observed as $gmv_{i}\mathcal{I}(\mathbf{x}_{i})$.  Ideally, we would like to choose the actions so that the total rewards are maximized. 

\subsubsection{Definition 2.4.2 (Rewards in DNN-MAB)}
The expected rewards are defined as the total GMV generated from the listed items that users place orders on.  In general it shall be translated into the company's operating revenue:  
\begin{eqnarray}
\mathcal{R}e & = & \sum^K_{i=1} gmv_{ji}\mathcal{I}(\mathbf{x}_{ji})
\end{eqnarray}
where $\{\mathbf{x}_{j1}, ~\mathbf{x}_{j2},~\ldots,~\mathbf{x}_{jK}\}$ is the ranked sublist from $\mathcal{T}_N$ which is co-decided by both DNN and MAB.  

The revised Thompson sampling algorithm is triggered after \emph{learning-to-rank} DNN \emph{pre}-ranker and we describe it in Algo.\ref{algo:DNN-MAB-TS}.  The main idea of Algo.\ref{algo:DNN-MAB-TS} is to take the \emph{pre}-ranker's output as the initial static ranking and finetune the local orders via users' online feedbacks so to reflect the current user purchasing intent in the final ranking.  Some important parameters are highlighted as follows: ${\scriptstyle\mathrm{SCALE}}$ is to adjust the intensity from negative feedbacks, alleviating the potential issue with the treatment that most items of \emph{no}-actions are seen as negative; $\theta_1,~\theta_2,~\theta_3$ are to control the weights for how much the \emph{pre}-ranking scores are to be changed; $\mathcal{U}_c$ is the set of items from arm $c$ that have not yet been selected; $\mathcal{E}_c$ is the set of items from arm $c$ that are presented but not clicked by users during \emph{post}-ranking; $\mathcal{A}_c$ is the set of items that are presented and clicked; $|\cdot|_0$ is the cardinality computation.   

\begin{algorithm}
\caption{\emph{post}-ranker: DNN-MAB Thompson sampling}\label{algo:DNN-MAB-TS}
\begin{algorithmic}[1]
\Procedure{INITIALIZATION}{}
\For{each $\left<\mathbf{x},y\right> \in \mathcal{T}_N$}
\State for arm $c$ such that $\mathbf{x} \in c$ 
\State $\alpha_c~=~\alpha_c + y$
\State $\beta_c~=~\beta_c + (1 - y)$
\State $\mathcal{U}_c \leftarrow \left<\mathbf{x},y\right>$
\State $avg_{c}~=~\alpha_{c} / |\mathcal{U}_c|_0$
\EndFor
\EndProcedure
\Procedure{At round-$i$ MAB ranking}{}
\State{\small PULLING ARMS:}
\For{each arm $c \in \mathcal{C}_M$}
\State draw sample $r~\sim~\mathrm{beta}(\alpha_c,~\beta_c)$
\State update all $y~=~y *(1 + {r} / \theta_1)$ for $\left<\mathbf{x},y\right>\in c$
\EndFor
\State pick $\left<\mathbf{x}_i, y_i\right> = \mathrm{arg}\max_{(y, r_{c})} \{\left<\mathbf{x},y\right>\in c\}$ 
\State $\mathcal{U}_c = \mathcal{U}_{c_i} - \left<\mathbf{x}_i,y_i\right>$
\State
\State{\small FEEDBACK:}
\If {$\left<\mathbf{x}_i, y_i\right>$ is exposed but not clicked}
\State $\mathcal{E}_c \leftarrow \left<\mathbf{x}_i, y_i\right>$
\State $\beta_{c_i}~=~\beta_{c_i} + (1 - avg_{c_i})*(1-\exp(-\frac{|\mathcal{E}_{c_i}|_0}{\mathrm{SCALE}}))* \theta_2$
\EndIf
\If {$\left<\mathbf{x}_i, y_i\right>$ is exposed and clicked}
\State $\mathcal{A}_{c_i} \leftarrow \left<\mathbf{x}_i, y_i\right>$
\State $\alpha_{c_i} ~=~\alpha_{c_i} + avg_{c_i}*(\frac{|\mathcal{A}_{c_i}|_0}{|\mathcal{E}_{c_i}|_0}) * \theta_3$
\EndIf
\EndProcedure
\end{algorithmic}
\end{algorithm}

At round $i$, DNN-MAB Thompson sampling randomly draws $M$ samples $\{r\}_M$ based upon the $M$ beta distributions estimated, and then selects the arm with the $\max$ $r_i$ and the item in that arm containing the $\max$ adjusted score $y_i$.  If it is clicked after exposure, the algorithm updates the beta distribution parameter $\alpha_{c_i}$ in arm $c_i$.  Otherwise the algorithm updates the beta distribution parameter $\beta_{c_i}$ in arm $c_i$.  

\section{EXPERIMENTAL EVALUATION}
\label{sec:experiment}
\subsection{Case study}
Before the real system online a/b test discussion, we first walk through a simple case study to evaluate different bandit algorithms' performance under our use cases.  We picked three state-of-the-art bandit algorithms: $\epsilon$-greedy\cite{watkins1989learning}, Upper Confidence Bound or UCB\cite{auer2002finite}, and Thompson sampling.  Specifically, we simulate two versions of Thompson sampling: 1. the revised Thompson sampling with the specially designed initialization (\emph{revised}-Thompson) (Algo.\ref{algo:DNN-MAB-TS}); 2. the normal Thompson sampling (\emph{normal}-Thompson).  The random selection is also performed serving as a naive baseline. 

In our simulation, we design $M=5$ arms and simply set each item's reward as $1$ if the user clicks, $0$ if the user does not click.  The way we define the `click' action is by presetting a thresholding probability $f_\mathrm{threshold}$, once the item is selected, we randomly generate another probability $f_\mathrm{item}$ via the real unknown beta distribution.  If $f_\mathrm{item} \geq f_\mathrm{threshold}$, we assume as the `click' action happens, otherwise we assume the customer is not interested in the item selected at this round.

\begin{figure}
\hskip -.5cm \includegraphics[width=4.5cm,height=3.5cm]{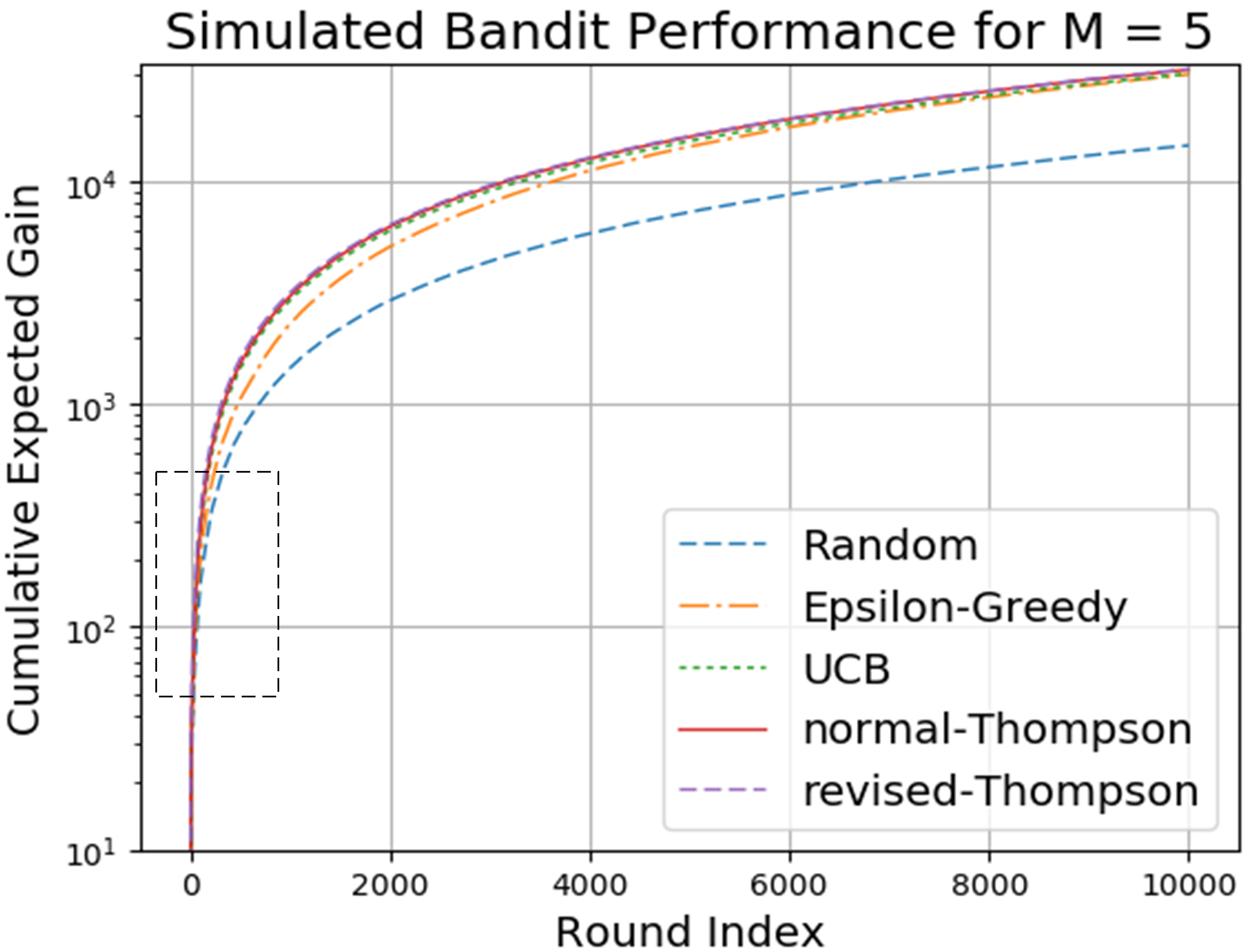}
\hskip -.1cm \includegraphics[width=4.5cm]{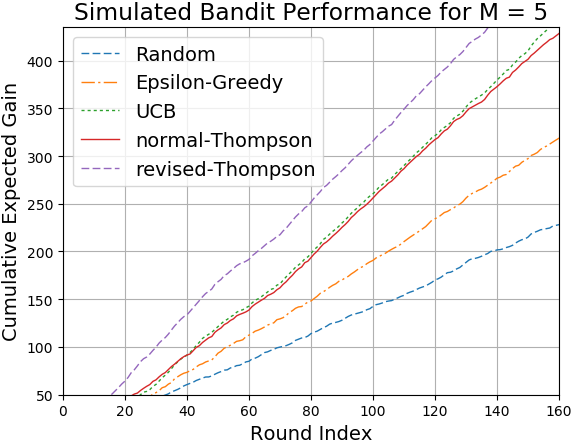}
\caption{multi-armed bandit algorithms rewards simulations}
\label{graph:simulation-mab-gain}
\end{figure}

\begin{figure}
\hskip -.5cm \includegraphics[width=4.5cm,height=3.5cm]{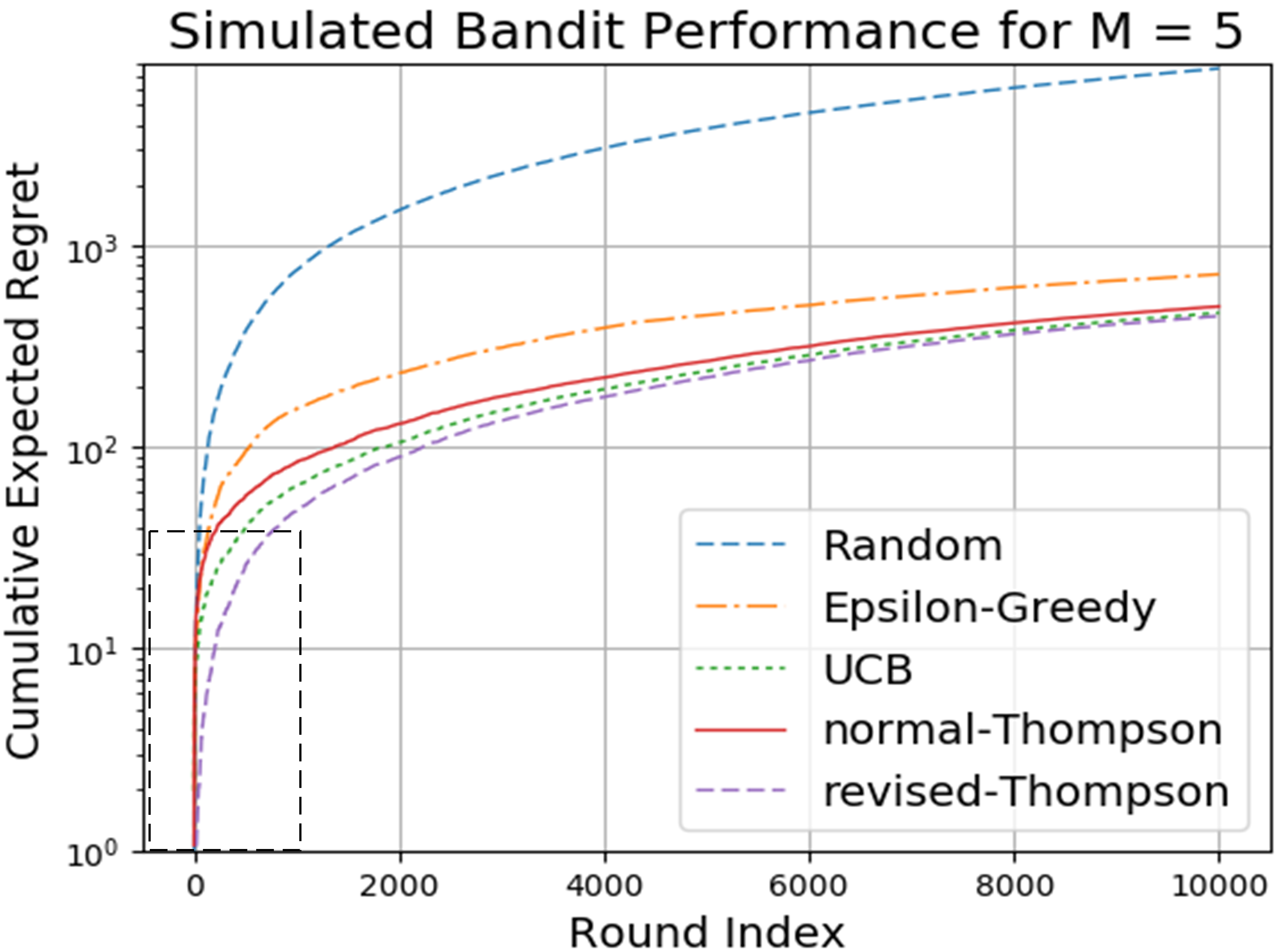}
\hskip -.1cm \includegraphics[width=4.5cm,height=3.5cm]{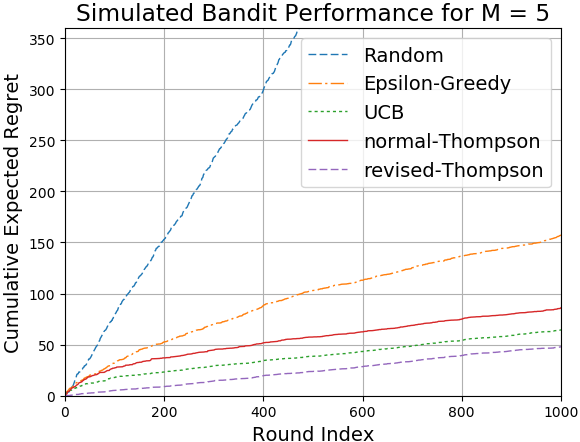}
\caption{multi-armed bandit algorithms regrets simulations}
\label{graph:simulation-mab-loss}
\end{figure}

We perform the simulation $10$ times and each simulation keeps running over $10,000$ rounds.  The average performance is shown in Fig.\ref{graph:simulation-mab-gain} \& \ref{graph:simulation-mab-loss}.  The left subfigures of Fig.\ref{graph:simulation-mab-gain} \& \ref{graph:simulation-mab-loss} are about the cumulative gains / regrets and the right ones are their zoom-ins.  As shown, $\epsilon$-greedy remains sub-optimal regarding with both rewards and regrets, UCB and \emph{normal}-Thompson perform almost equally well, and the \emph{revised}-Thompson performs the best by beating UCB and \emph{normal}-Thompson with faster convergence.  This is due to the fact that the \emph{revised}-Thompson's initialization phase personalizes the arms based upon the user information.  Hence, \emph{revised}-Thompson could converge in less steps relative to other standard MAB algorithms.  The random selection no-surpisingly performs the worst among the five.  Implementation-wise, the \emph{revised}-Thompson is also straightforward and the overall system latency remains low (reported in Sec.\ref{subsection:system-specifications}).  With the above arguments, the \emph{revised}-Thompson becomes the choice of our \emph{post}-ranker module.  


\subsection{Experiment setup}
JD processes billions of requests in a daily basis, any new models about to launch have to be evaluated by JD's online testing platform.   It divides the real traffics into $10$ buckets, each bucket gets about $8\%$ of the total traffics, and the remaining $20\%$ is held by the control bucket.

We deploy the proposed dynamic ranking paradigm on this platform for $7$ days, and track following metrics: GMV, order numbers, overall (Eq.\ref{eqn:def-ndcg}) and page-wise (Eq.\ref{eqn:def-page-wise-dcg}) discounted cumulative gains ($dcg$).  

\begin{eqnarray}
dcg_{p,\mathrm{page}-k} ~=~ \sum^{i=p}_{i=1, k\in \mathrm{page}}\cfrac{gmv_{ki}\mathcal{I}(\mathbf{x}_{ki})}{\log_2 (i + 1)} 
\label{eqn:def-page-wise-dcg}
\end{eqnarray}

Since the item lists are presented to users in a page-wise fashion and each page contains $4$ - $20$ items, page-wise ${dcg}_p$ is a perfect metric for evaluating how is the \emph{revised}-Thompson module functioning in the application and how much gains we observed should be credited to it (we use $p~=~8$ in the evaluation). 

\subsection{Production performance}
\begin{figure}
\centering
\includegraphics[width=8.5cm]{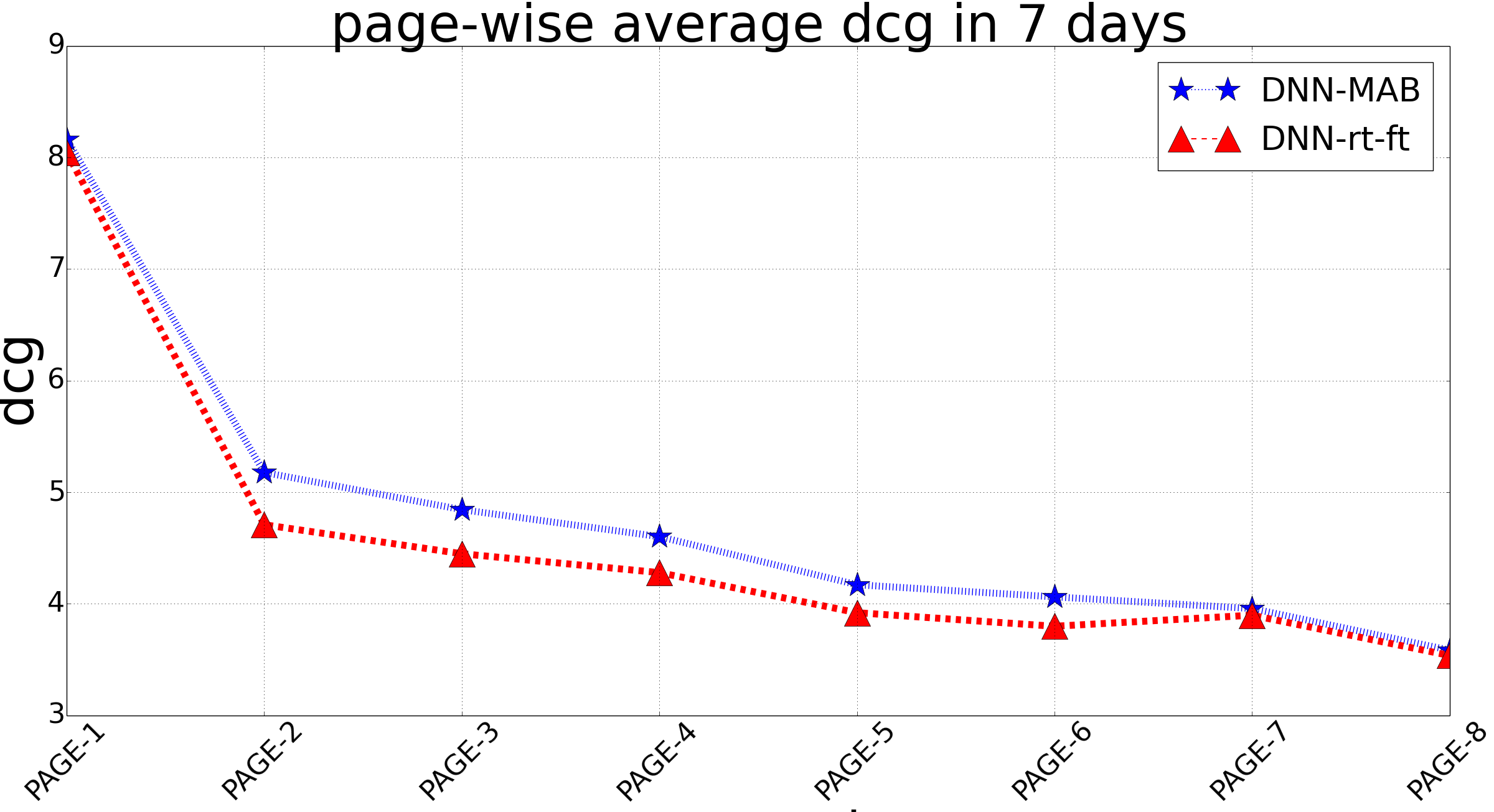}
\caption{page-wise dcg gain: DNN-MAB v.s. DNN-\emph{rt-ft}}
\label{graph:page-wise-dcg-comp}
\end{figure}
We report the performance between DNN-MAB and DNN-\emph{rt-ft} as the baseline in Tab.\ref{table:gmv_order_cn}.  DNN-\emph{rt-ft} is to utilize the DNN \emph{pre}-ranker by taking both offline users' feedbacks, online browsing and click signals as features for training a near-line model that is better than models taking offline signals only.  In average we see DNN-MAB's daily GMV has increased $\mathbf{16.69\%}$ over DNN-\emph{rt-ft}.   Any performance gains that are greater than $1.0\%$ over $7$ days are considered statistically significant in the real production system.  DNN-MAB paradigm has clearly proved its superiority against the current production DNN-\emph{rt-ft} ranking strategy.  
To emphasize the importance of the parameter initialization and the feeback revision in Thompson sampling, we also report DNN + \emph{normal}-Thompson in Tab.\ref{table:gmv_order_cn_no_ini}.  Due to the space limitation, we do not go to the very details but simply put our conclusion that DNN + \emph{normal}-Thompson in general will not beat DNN-\emph{rt-ft} because it can not learn users online behaviors quickly enough to improve the ranking quality.  
\begin{figure}
\centering
\includegraphics[width=8.5cm]{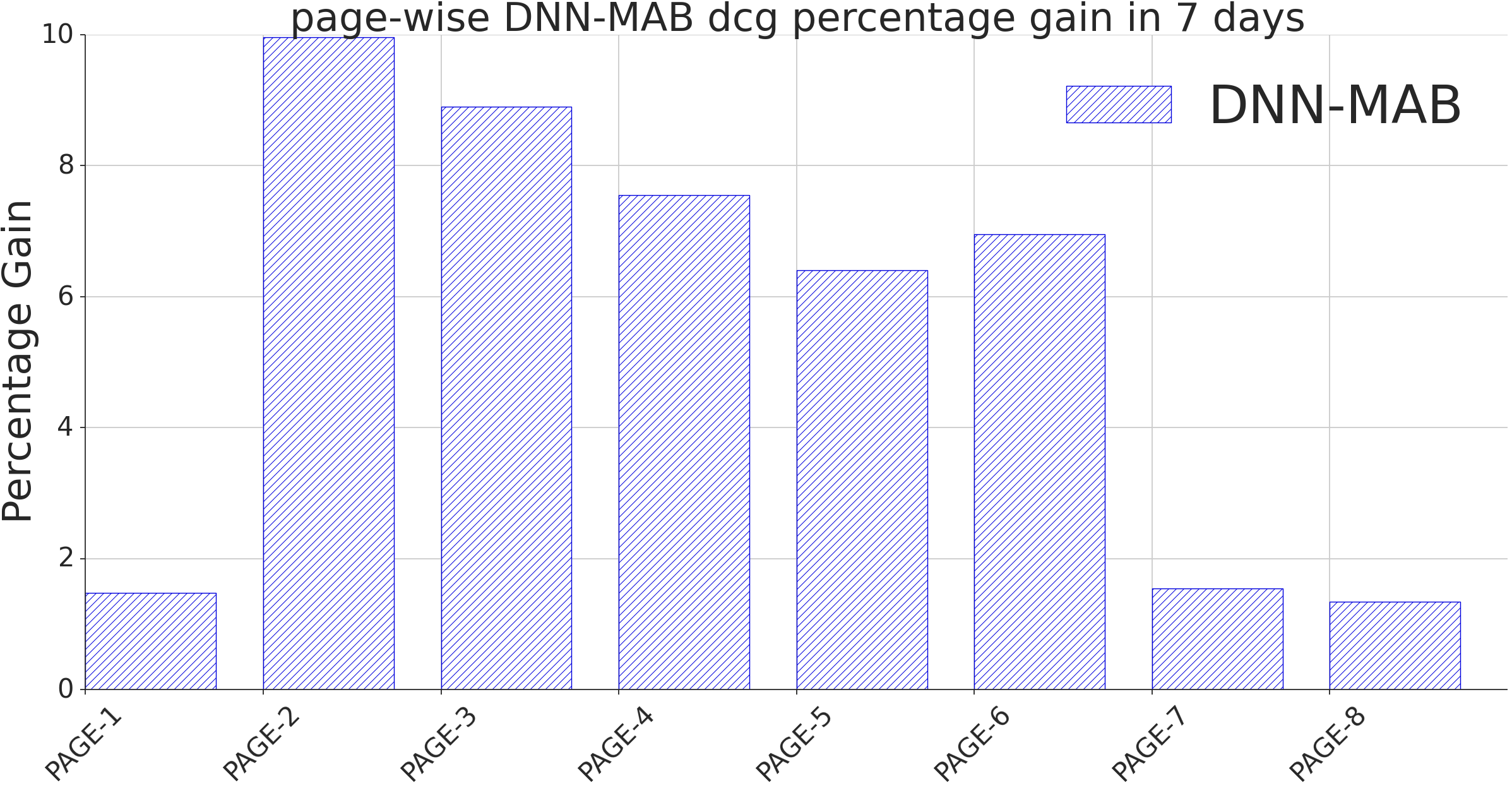}
\caption{7-day page-wise DNN-MAB dcg percentage gain}
\label{graph:page-wise-dcg-gain}
\end{figure}
We also report the overall $dcg$ gains in Tab.\ref{table:time_avg_dcg} and page-wise $dcg$ gains in Tab.\ref{table:page_avg_dcg}.  At the first glance, it seems that DNN-MAB beats the production baseline consistently in terms of overall $dcg$ as well as page-wise $dcg$.  By taking a closer look at the page-wise $dcg$ comparison (Fig.\ref{graph:page-wise-dcg-comp}) and the MAB-DNN page-wise $dcg$ gains in percentage (Fig.\ref{graph:page-wise-dcg-gain}), we find that the \emph{revised}-Thompson is able to effectively learn the users' intent.  Due to the fact that the \emph{revised}-Thompson takes each user's recent behaviors for the personalized initialization and keeps tracking the real-time user browsing signals for the dynamic ranking adjustment.  Although the page-wise percentage gains are not quite visible at the page-1 ($+1.47\%$), the dynamic ranking performances are maximized at page-2 ($+9.96\%$) and page-3 ($+8.90\%$), and then deminish along with users browsing more and more pages.  In the end DNN-MAB and DNN-\emph{rt-ft} both end up with similar page-wise performances at page-7 ($+1.54\%$) and page-8 ($+1.34\%$).   
\begin{table}
\hskip -.25cm
\begin{tabular}{ |p{0.4cm}|p{0.6cm}|p{0.6cm}|p{0.6cm}|p{0.6cm}|}
\hline
{\scriptsize\bf Date} & {\scriptsize Day1} & {\scriptsize Day2} & {\scriptsize Day3} & {\scriptsize Day4} \\
\hline
{\scriptsize\bf GMV} &{\scriptsize \hskip -.3cm +22.87\%} &{\scriptsize \hskip -.3cm +45.45\%} &{\scriptsize \hskip -.3cm +20.20\%} &{\scriptsize \hskip -.25cm +2.73\%} \\
{\scriptsize\bf Orders} &{\scriptsize \hskip -.25cm -2.14\%} &{\scriptsize \hskip -.25cm -1.57\%} &{\scriptsize \hskip -.25cm +5.18\%} &{\scriptsize \hskip -.25cm +0.42\%} \\
\hline
{\scriptsize\bf Date} & {\scriptsize Day5} & {\scriptsize Day6} & {\scriptsize Day7} & {\scriptsize\bf Summary} \\
\hline
{\scriptsize\bf GMV} & {\scriptsize \hskip -.25cm +0.91\%} & {\scriptsize \hskip -.3cm +23.15\%} & {\scriptsize \hskip -.25cm +1.50\%} & {\scriptsize\bf \hskip -.35cm +16.69\%} \\
{\scriptsize\bf Orders} & {\scriptsize \hskip -.25cm -2.79\%} & {\scriptsize \hskip -.25cm +4.19\%} & {\scriptsize \hskip -.25cm +2.20\%} & {\scriptsize\bf \hskip -.3cm +0.78\%} \\
\hline
\end{tabular}
\caption{GMV and orders gain / loss for DNN-MAB}
\label{table:gmv_order_cn}
\end{table}

\begin{table}
\hskip -.25cm
\begin{tabular}{ |p{0.4cm}|p{0.6cm}|p{0.6cm}|p{0.6cm}|p{0.6cm}|}
\hline
{\scriptsize\bf Date} & {\scriptsize Day1} & {\scriptsize Day2} & {\scriptsize Day3} & {\scriptsize Day4} \\
\hline
{\scriptsize\bf GMV} &{\scriptsize \hskip -.3cm -12.08\%} &{\scriptsize \hskip -.3cm -9.33\%} &{\scriptsize \hskip -.3cm -4.74\%} &{\scriptsize \hskip -.25cm -3.24\%} \\
{\scriptsize\bf Orders} &{\scriptsize \hskip -.25cm 0.30\%} &{\scriptsize \hskip -.25cm -4.72\%} &{\scriptsize \hskip -.25cm -1.34\%} &{\scriptsize \hskip -.25cm -0.67\%} \\
\hline
{\scriptsize\bf Date} & {\scriptsize Day5} & {\scriptsize Day6} & {\scriptsize Day7} & {\scriptsize\bf Summary} \\
\hline
{\scriptsize\bf GMV} & {\scriptsize \hskip -.25cm -18.31\%} & {\scriptsize \hskip -.3cm -7.49\%} & {\scriptsize \hskip -.25cm -1.43\%} & {\scriptsize\bf \hskip -.35cm -8.08\%} \\
{\scriptsize\bf Orders} & {\scriptsize \hskip -.25cm -10.67\%} & {\scriptsize \hskip -.25cm -4.69\%} & {\scriptsize \hskip -.25cm -0.81\%} & {\scriptsize\bf \hskip -.3cm -3.23\%} \\
\hline
\end{tabular}
\caption{GMV and orders gain / loss for DNN + \emph{normal}-Thompson}
\label{table:gmv_order_cn_no_ini}
\end{table}

\subsection{System specifications}
\label{subsection:system-specifications}
Our current DNN-MAB ranking system is maintained by hundreds of Linux servers\footnote{We could not release this piece of information regarding with the exact number of operating servers due to the company confidentiality.}, the \emph{qps} (query per second) is $32$ in average (peak at $52$), and the overall recomendation end-to-end response latency is within $50.0$ \emph{milli}-seconds (including both retrieval and ranking phases).   

\begin{table}
\centering
\begin{tabular}{ |p{0.8cm}|p{1.0cm}|p{1.0cm}|p{0.8cm}|  }
\hline
{\bf Date} & {\hskip -.3cm \bf\scriptsize DNN-MAB} & {\hskip -.3cm\bf\scriptsize DNN-\emph{rt-ft}} & {\bf\small Gain} \\
\hline
Day1 & 5.470 & 5.180 & +5.60\%\\
Day2 & 5.303 & 4.811 & +10.23\%\\
Day3 & 5.434 & 5.281 & +2.90\%\\
Day4 & 5.443 & 5.340 & +1.93\%\\
Day5 & 4.865 & 4.789 & +1.59\%\\
Day6 & 5.873 & 5.491 & +6.96\%\\
Day7 & 7.045 & 6.884 & +2.34\%\\
\hline
{\bf Average} & 5.633 & 5.397 & {\bf+4.37\%}\\
\hline
\end{tabular}
\caption{dcg online a/b test: DNN-MAB v.s. DNN-\emph{rt-ft}}
\label{table:time_avg_dcg}
\end{table}

\begin{table}
\centering
\begin{tabular}{ |p{1.0cm}|p{1.0cm}|p{1.0cm}|p{0.8cm}|  }
\hline
{\bf Page} & {\bf\hskip -.3cm \scriptsize DNN-MAB} & {\hskip -.3cm\bf\scriptsize DNN-\emph{rt-ft}} & {\bf\small Gain} \\
\hline
Page-0 & 8.164 & 8.046 & +1.47\%\\
Page-1 & 5.177 & 4.708 & +9.96\%\\
Page-2 & 4.844 & 4.448 & +8.90\%\\
Page-3 & 4.602 & 4.279 & +7.55\%\\
Page-4 & 4.171 & 3.920 & +6.40\%\\
Page-5 & 4.062 & 3.798 & +6.95\%\\
Page-6 & 3.957 & 3.897 & +1.54\%\\
Page-7 & 3.584 & 3.536 & +1.34\%\\
\hline
\end{tabular}
\caption{page-wise dcg a/b test: DNN-MAB v.s. DNN-\emph{rt-ft}}
\label{table:page_avg_dcg}
\end{table}

\section{RELATED WORK}
\label{sec:related}
\subsection{Recommender system}
The recommender system is the key to the success of E-Commerce websites as well as other indexing service providers, such as Alibaba, Ebay, Google, Baidu, Youtube, etc.  Efforts from different parties regarding with how the recommender systems should be designed include \cite{linden2003amazon,davidson2010youtube,schafer2001commerce,sarwar2000analysis}.  There are in general two streams of works in the recommender system researches: content-based approaches and item-based approaches.  Item-based approaches represent users and items as a huge $M$ by $N$ matrix and focus on learning the underlying relations between items.  Works of item-based approaches such like \cite{sarwar2001item,rendle2010factorizing}
  have all received enormous success.  Yet item-based approaches suffer from issues like \emph{cold-start}, scalability and plasticity, etc.  Content based approaches treat the problem as the query-indexing problem, which in general, scales better and performs well for cases that users do not have too many previous actions in records but it tends to have query generalization issues for users with many behavior histories.  \cite{burke2002hybrid,lops2011content} both provide thorough surveys about this topic and readers should refer to them for in depth details. 

\subsection{\emph{learning-to-rank} via deep neural network}
\emph{Learning-to-rank}, emerged from late 90s, has always been an interesting research topic in information retrieval.  Approaches for solving this problem could be summarized into two main threads depending on the loss funtions that different approaches utilize: the pairwise loss and the listwise loss.  In pairwise approaches, it has been formulated as a classification problem: item pairs are generated by picking up samples from positive and negative classes, the goal for \emph{learning-to-rank} models is to correctly categorize item pairs into the binary classes, so that the loss defined is minimized.  Research works in this thread include \cite{freund2003efficient,cao2006adapting}.  Listwise approaches, on the other hand, formulate the ranking problem as a classification problem on permulations.  The loss will only be minimized if the whole list is perfectly ranked.  Successful listwise approaches include ListNet\cite{cao2007learning} and RankCosine\cite{qin2008query}.  For more in-depth discussions regarding with \emph{learning-to-rank}, please refer to \cite{liu2009learning}. 

With the growth in popularity of deep learning, people start to think of using different deep learning structures to tackle the \emph{learning-to-rank} problem.  Maybe the works that are most similar to our \emph{pre}-ranker could be \cite{severyn2015learning,kalchbrenner2014convolutional,kim2014convolutional}.  They utilized convolutional deep neural network for ranking texts in natural languarage processing problems.  

\subsection{Contextual multi-armed bandit problems}
Multi-armed bandit problem has been well studied and discussed in literatures such like \cite{lai1985asymptotically,even2006action,auer2002finite}  The basic setup for MAB is to select $K$ items from $M$ arms consecutively with feedbacks so the total expected regrets are minimized.  Thompson sampling, dated back from \cite{thompson1933likelihood}, albeit its simplicity, has been proved quite efficient regarding with productional performance \cite{tang2013automatic,chapelle2011empirical}
  and suprisingly straightforward to implement.  Other works regarding with Thompson sampling models include \cite{scott2010modern}. 

\section{CONCLUSION}
\label{sec:conclusion}
We proposed a dynamic ranking framework called DNN-MAB which is composed of \emph{learning-to-rank} DNN \emph{pre}-ranker and \emph{revised}-Thompson \emph{post}-ranker.  This effective ranking paradigm takes into consideration of both the user and the item information so that the DNN could reach the decent static ranking performance.  Meanwhile, by tracking real time user feedbacks, the \emph{revised}-Thompson sampling is able to adjust the \emph{pre}-rankings that futher boosts the objective metrics.  To our knowledge, such a ranking framework has not been discussed in previous researches.  Real production tests show that both GMVs and $dcg$s have been significantly improved.  However, for the sake of model simplicity, we have not paid too much attention to the position-\emph{bias} which is one important factor that affects the ranking performance \cite{radlinski2008learning} in the \emph{learning-to-rank} DNN \emph{pre}-ranker, since bringing in the listwise loss would introduce some scalability issues in our use cases.  Meanwhile we have not optimized the proposed paradigm regarding with other user behaviors such as `clicks', `orders', etc. either (we do observe negative moves in several days regarding with order numbers, which is reported in Tab.\ref{table:gmv_order_cn}).  Which said, how to optimize multiple \emph{\small KPI}s simultaniously still remains as a big challenge.  We plan to further improve our ranking models along with the these research paths in the future.

\subsection*{Acknowledgements} 
We are thankful to Dali Yang, Huisi Ou, Sulong Xu, Jincheng Wang, Lu Bai, Lixing Bo as well as anonymous reviewers for their helpful comments.  This research has been supported in part by JD Business Growth BU and JD Santa Clara Research Center.

\bibliographystyle{named}
\bibliography{main}

\end{document}